# Artificial Intelligence for Conflict Management

E. Habtemariam*
T. Marwala*
*School of Electrical and Information Engineering
University of the Witwatersrand
Johannesburg, South Africa
E-mail: e.habtemariam@ee.wits.ac.za
tmarwala@yahoo.com

M. Lagazio
Department of Politics and International Relations
University of Kent
Canterbury, Kent
E-mail: M.Lagazio@kent.ac.uk

**Abstract**

Militarised conflict is one of the risks that have a significant impact on society. Militarised Interstate Dispute (MID) is defined as an outcome of interstate interactions, which result on either peace or conflict. Effective prediction of the possibility of conflict between states is an important decision support tool for policy makers. In a previous research, neural networks (NNs) have been implemented to predict the MID. Support Vector Machines (SVMs) have proven to be very good prediction techniques and are introduced for the prediction of MIDs in this study and compared to neural networks. The results show that SVMs predict MID better than NNs while NNs give more consistent and easy to interpret sensitivity analysis than SVMs.

## I. INTRODUCTION

Militarised Interstate Disputes (MID) as defined by (Gochman and Maoz, 1984) and by (Marwala and Lagazio 2004) refers to the threat of using military force between sovereign states in an explicit way. In other words, MID is a state that results from interactions between two states, which can be either peace or conflict. These interactions are expressed in the form of dyadic attributes which are two states' parameters considered to influence the probability of military conflict (Beck, King and Zeng 2000). In this study, seven dyadic variables are employed to predict MID outcome.

Interstate conflict is a complex phenomenon that encompasses non-linear pattern of interactions (Beck, King and Zeng, 2000; Lagazio and Russett, 2003; Marwala and Lagazio, 2004). Various efforts have and still are underway to improve the MID data, the underlying theory and the statistical modelling techniques of interstate conflict (Beck, King and Zeng, 2000.) Previously, linear statistical methods were used for quantitative analysis of conflicts, which were far from satisfactory. The results obtained showed high variance, which make them difficult to be reliable (Beck, King and Zeng, 2000). The results have to be taken cautiously and their interpretations require prior knowledge of the problem domain. This makes it inevitable to look for techniques other than the traditional statistical methods to do quantitative analysis of international conflicts. Artificial intelligence techniques have proved to be very good in modelling complex and nonlinear problems without any *a priori* constraints about the underlying functions assumed to govern the distribution of MID data (Beck, King and Zeng, 2000). It then makes sense to model interstate conflicts using artificial intelligence techniques.

Neural networks have previously been used by (Marwala and Lagazio, 2004; Beck, King and Zeng, 2000; Lagazio and Russett, 2003) to model MID. In this paper, two artificial intelligence techniques, neural networks and support vector machines, are used for the same purpose and their results were compared. These two techniques have been compared for other applications (e.g. for text texture verification (Chen and J. Odobez, 2002)] and for option pricing (Pires and

Marwala, 2004), etc. Their findings show that SVMs have outperformed NNs. These results motivate the use of SVMs to model interstate conflict.

II. BACKGROUND

*A. Learning machines*

There are different types of learning machines used for different kinds of purposes. The two major applications of learning machines are for classification and regression. Learning problem for classification can be defined as finding a rule that assigns objects into different classes (Müller et al, 2001). The rule that governs the classification is devised based on an acquired knowledge about the object. The process of knowledge acquisition is called training. In this paper, we look at two different types of learning machines for the purpose of predicting MID. Both techniques learn an underlying pattern based on training MID data to predict previously unseen (test) MID data.

*B. Artificial Neural Networks (NNs)*

Neural network is a processor that resembles the brain in its ability to acquire knowledge from its environment and store the information in some synaptic weights (Haykin, 1999). It is composed of simple neurons that are capable of processing information in a massively parallel fashion and can also make generalisations once they are trained using training data. This property gives neural networks the ability to solve complex non-linear problems (Haykin, 1999).

The most widely used neural network is a feed-forward network with two layers of adaptive weights (Bishop, 1995). As it is shown in figure 1, it has input, hidden and output layers. The input layer represents the independent variables while the hidden and output layers represent the latent and dependent variables, respectively (Zeng, 1999). Feed-forward neural networks provide a framework to represent a nonlinear functional mapping of a set of *d* input variables $x_i$, $i = 1, ...,d$ into a set of *c* output variables $y_j$, $j = 1, ..., c$ ( Bishop, 1995).

The relationship between the input and output units of the neural network is represented by the following function (Bishop, 1995; Marwala and Lagazio, 2004)]:

$$y_k = f_{outer}\left(\sum_{j=1}^{M} W_{kj}^{(2)} f_{inner}\left(\sum_{i=1}^{d} W_{ji}^{(1)} x_i + W_{j0}^{(1)}\right) + W_{k0}^{(2)}\right) \qquad (1)$$

where $W_{ji}^{(1)}$ and $W_{kj}^{(2)}$ are the first and second layer weights going from input *i* to hidden unit *j* and hidden unit *j* to output unit *k*, respectively. *M* is number of the hidden units, *d* is number of input units, while $W_{j0}^{(1)}$ and $W_{k0}^{(2)}$ represent biases of the hidden and output units, respectively. The output activation function is given by $f_{outer}$ while for the hidden unit it is $f_{inner}$.

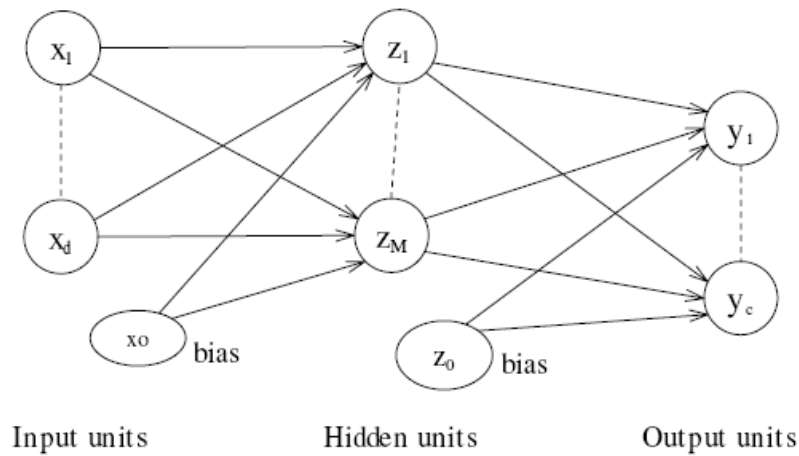

Fig. 1. A feed-forward network with two layers of adaptive weights (Marwala and Lagazio, 2004)

*C. Support Vector Machines (SVMs)*

According to (Müller et al, 2001)], the classification problem can be formally stated as estimating a function $f: R^N \rightarrow \{-1, 1\}$ based on an input-output training data generated from an independently, identically distributed unknown probability distribution P(**x**,y) such that *f* will be able to classify previously unseen (**x**,y) pairs. The best such function is the one that minimises the expected error (risk) which is given by

$$R[f] = \int l(f(x), y) dP(x, y) \qquad (2)$$

where *l* represents a loss function (Müller et al, 2001). Since the underlying probability distribution *P* is unknown, equation (2) cannot be solved directly. The best we can do is to find an upper bound for the risk function (Vapnik, 1995), (Müller et al, 2001) which is given by

$$R[f] = R[f]_{emp} + \sqrt{\frac{h\left(\ln\frac{2n}{h} + 1\right) - \ln\left(\frac{\delta}{4}\right)}{n}} \qquad (3)$$

where $h \in \mathbb{N}^+$ is the Vapnik-Chervonenkis *(VC)* dimension of $f \in F$ and $\delta > 0$ holds true for all. The *VC* dimension of a function class *F* is defined as the largest number *h* of points that can be separated in all possible ways using functions of that class (Vapnik, 1995). The empirical error $R[f]_{emp}$ is a training error given by

$$R[f]_{emp} = \frac{1}{n} \sum_{i+1}^{n} l(f(x_i), y_i) \qquad (4)$$

Assuming that the training sample is linearly separable by a hyperplane of the form

$$f(x) = (w.x) + b \qquad (5)$$

where **w** is an adjustable weight vector and **b** is an offset, the classification problem looks like figure 2 (Müller et al, 2001)

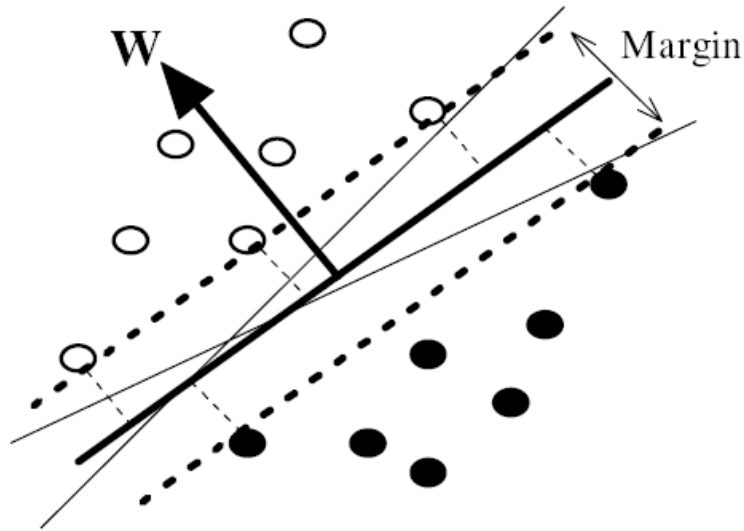

Fig. 2. A linear SVM classifier and margins: A linear classifier is defined by a hyperplane's normal vector **w** and an offset *b*, i.e. the decision boundary is {**x**|**w.x** + *b* =0} (thick line). Each of the two half spaces defined by this hyperplane corresponds to one class, i.e. *f(x) = sign((**w.x**) + b)*. (Müller et al, 2001)

The goal of the learning algorithm as proposed by (Vapnik and Lerner, 1963), is to find the hyperplane with maximum margin of separation from the class of separating hyperplanes. But since real-world data often exhibit complex properties, which cannot be separated linearly, more complex classifiers are required. In order to avoid the complexity of the nonlinear classifiers, the idea of linear classifiers in a feature space comes into place. Support vector machines try to find a linear separating hyperplane by first mapping the input space into a higher dimensional feature space $\mathcal{F}$. This implies each training example $x_i$ is substituted with $\Phi(x_i)$

$$Y_i((w \cdot \Phi(x_i) + b), \quad i=1,2,\ldots n \tag{6}$$

The *VC* dimension $h$ in the feature space $\mathcal{F}$ is bounded according to $h \leq \|W\|^2 R^2 + 1$ where $R$ is the radius of the smallest sphere around the training data (Müller et al, 2001). Hence minimising the expected risk is stated as an optimisation problem

$$\min_{w,b} \quad \frac{1}{2}\|W\|^2 \qquad (7)$$

However, assuming that we can only access the feature space using only dot products, (7) is transformed into a dual optimisation problem by introducing Lagrangian multipliers $\alpha_i$, $i = 1,2,..., n$ and doing some minimisation, maximisation and saddle point property of the optimal point (Burges, 1998; Müller et al, 2001; Schölkopf and Smola, 2003), the problem becomes

$$\text{Max } \alpha \quad \sum_{i=1}^{n}\alpha_i - \frac{1}{2}\sum_{i,j=1}^{n}\alpha_i\alpha_j y_i y_j k(x_i, x_j) \qquad (8)$$

$$\text{subject to} \quad \alpha_i \geq 0, i = 1,...,n$$

$$\sum_{i=1}^{n}\alpha_i y_i = 0$$

The Lagrangian coefficients $\alpha_i$'s, are obtained by solving (8) which in turn is used to solve **w** to give the non-linear decision function (Müller et al, 2001):

$$\begin{aligned}f(x) &= \text{sgn}\left(\sum_{i=1}^{n}y_i\alpha_i(\Phi(x).\Phi(x_i)) + b\right)\\ &= \text{sign}\left(\sum_{i=1}^{n}y_i\alpha_i k(x, x_i) + b\right)\end{aligned} \qquad (9)$$

In the case when the data is not linearly separable, a slack variable $\xi_i$, $i = 1, ..., n$ is introduced to relax the constraints of the margin as

$$y_i((w,\Phi(x_i)) + b) \geq 1 - \xi_i, \; \xi_i \geq 0, i = 1,...,n \qquad (10)$$

A trade off is made between the VC dimension and the complexity term of (3) which gives the optimisation problem

$$\min_{\mathbf{w}, b, \xi} \quad \frac{1}{2}\|W\|^2 + C\sum_{i=1}^{n} \xi_i \tag{11}$$

where $C > 0$ is a regularisation constant that determines the above-mentioned trade-off. The dual optimisation problem is then given by (Müller et al, 2001):

$$\max_{\alpha} \sum_{i=1}^{n} \alpha_i - \frac{1}{2}\sum_{i,j=1}^{n} \alpha_i \alpha_j y_i y_j k(\mathbf{x}_i, \mathbf{x}_j) \tag{12}$$

$$\text{subject to } 0 \leq \alpha_i \leq C, i = 1, ..., n$$

$$\sum_{i=1}^{n} \alpha_i y_i = 0$$

A Karush-Kuhn-Tucker (KKT) condition which says only the $\alpha_i$'s associated with the training values $\mathbf{x}_i$'s on or inside the margin area have non-zero values, is applied to the above optimisation problem to find the $\alpha_i$'s and the threshold variable $b$ reasonably and the decision function $f$ (Müller et al, 2001).

*D. Conflict Modelling*

Modelling international conflicts involve quantitative and empirical analysis based on existing dyadic information of countries. Dyad-year in our context refers to a pair of countries in a particular year. Political scientists apply dyadic parameters as a measure of the possibility that two countries will have a militarised conflict. Although extensive data collection efforts have been made, still a lot of research is underway to come up with satisfactory and reliable conflict models. One of the major reasons why conflict modelling is complex, according to (Beck, King, and Zeng,

2000), is that international conflict is a rare event and the processes that drive it vary for each incident. A small change made on the explanatory variables affects greatly the MID outcome. This makes modelling MID to be highly nonlinear, very interactive and context dependent. In modelling interstate conflict, (Marwala and Lagazio, 2004) used seven dyadic variables. They used MID data that came from Correlates of War (COW) which was compiled by (Russett and Oneal, 2001). Since the same variables, which are discussed in (Marwala and Lagazio, 2004; Lagazio and Russett, 2003), are used for this study, their brief description follows.

The variables are classified into *realist* and *kantianas* described by (Lagazio and Russett, 2003). The realist variables include *Allies, Contingency, Distance, Major power and Capability*. *Allies* is the measure of military alliance between the dyad countries. A value of 1 implies that there is an alliance of some kind between the two countries while a value of 0 indicates no alliance. *Contingency* measures whether the two countries have any common boundary. If they share a boundary its value becomes 1 and otherwise 0. *Distance* is the logarithm to the base 10 of the distance in Kilometres between the two states' capitals. *Major Power* is assigned 1 if one or both states in the dyad are major powers and otherwise 0. *Capability* is the logarithm to the base 10 of the ratio of the total population plus number of people in urban areas plus industrial energy consumption plus iron and steel production plus number of military personnel in active duty plus military expenditure in dollars in the last 5 years measured on stronger country to weak country.

The other variables used in this study, which are referred as Kantian, are *Democracy* and *Dependency*. *Democracy* is a scale in the range [-10, 10] where -10 denotes extreme autocracy and 10 for extreme democracy. The lowest value of the two countries is taken because it is assumed the less democratic state plays a determinant role for an occurrence of conflict (Oneal and Russett, 1999)]. *Dependency* is measured as the sum of a country's import and export with its partner divided by the Gross Domestic Product of the stronger country. It is a continuous variable that measures the level of economic interdependence (dyadic trade as a portion of a state's gross domestic product) of the less economically dependent state in the dyad.

## III. METHOD

### A. Neural networks

Neural network requires selecting the best architecture to give good classification results. The best combination of the number of hidden units, activation function, training algorithm and training cycles that can result in a network which is able to generalise the test data in the best possible way is searched for during the model selection process. This helps to avoid the risk of over/under training. A multi-layer perceptron (MLP) trained with the scaled conjugate gradient (SCG) method (Moller, 1993) neural network was used to classify the MID input data. Logistic and hyperbolic activation functions for the output and hidden layers respectively, $M = 10$ number of hidden units and 100 training cycles resulted in an optimal architecture.

### B. Support vector machines

SVM employs a method of mapping the input into a feature space of higher dimensionality and then finds a linear separating hyperplane with maximum margin of separation. There are various mapping or kernel functions in use, the most common of which are linear, polynomial, radial basis function (RBF) and sigmoid. The choice of a kernel function depends on the type of problem at hand and the RBF: $K(x_i, x_j) = \exp(-\gamma \| x_i - x_j \|^2, \gamma > 0$ can handle non-linear data better than the linear kernel function. Moreover, the polynomial kernel has a number of hyperparameters which influences the complexity of the model and some times its values may become infinity or zero as the degree becomes large. This makes RBF to be a common choice for use.

Similar to the neural network, SVM also requires selection of a model that gives an optimal result. The conducted experiments show that RBF gives best results for the classification of MID data in much less time. When RBF is used as the kernel function, there are two parameters that are required to be adjusted to give the best result. These are the penalty parameter of the error term $C$ and the $\gamma$ parameter of the RBF kernel function. Cross-validation and grid-search are two methods which are used to find an optimal model. For this experiment, a 10-fold cross-validation technique and a simple grid-search for the variables C and $\gamma$ were used. The parameters C = 1 and $\gamma$ = 16.75 gave the best results.

*C. MID data*

The data sets for the experiment, as discussed in the previous section, came from the Correlates of War (COW) that was compiled and used by Russett and Oneal (Russett and Oneal, 2001). It includes politically relevant dyads for the cold war and immediate post-cold war period (CW), from 1946 to 1992. As it is described in (Marwala and Lagazio, 2004; Lagazio and Russett, 2003; Oneal and Russett, 2001; Oneal and Russett, 1999), politically relevant dyads refer to all those which are contiguous and contain major power. Distant and weak dyads are omitted from the data set because it is less probable for them to have conflicts. Since the aim is to predict the onset of a conflict rather than its continuation, the dyads include only those with no disputes or only the initial year of the militarised conflict. The unit of analysis is a dyad-year. After the omission, a total of 27737 with 26845 peace and 892 conflict dyad-years were filtered out.

The dyadic data was classified into two sets, which are training and testing set. In their study (Lagazio and Russett, 2003), have given a detailed discussion on how the training set should be chosen. They have found out that a balanced set, equal number of conflict and peace dyads, gives best results as training set for the neural network. The same principle was adhered to in this study. The training set contains 1000 randomly chosen dyads, 500 from each group. The test set contains 26737 dyads of which 392 are conflict and 26345 non-conflict dyads.

## IV. RESULTS AND DISCUSSION

Neural network and support vector machine were employed to classify the MID data. The main focus of the result is to look at the percentage of correct MID prediction of the test data set by each technique. Table I depicts the confusion matrix of the results. Although NN performed as good as SVM in predicting true conflicts (true positives), this is achieved at the expense of reducing the number of correct peace (true negatives) prediction. SVM picked up the true conflicts (true positives) better than NN without effectively minimising the number of true peace (true negatives). SVM is able to pick 1450 more cases of the true peace (true negatives) than NN, which makes it a better choice than NN. Over all, SVM is able to predict peace and conflict with 79% and 75%, respectively. The corresponding results for NN are 74% and 76%, respectively for peace and conflict. The combined results of correct predictions are 79% for SVM and 74% for NN.

TABLE I

NN AND SVM CLASSIFICATION RESULTS

| Method | TC | FP | TP | FC |
|---|---|---|---|---|
| Neural Network | 297 | 95 | 19464 | 6881 |
| Support Vector Machine | 295 | 97 | 20914 | 5431 |

TC = true conflict (true positive), FC=false conflict (false positive), TP=true peace (true negative), and FP = false peace (false negative)

*A. Receiver operating characteristic (ROC) curve*

The receiver operating characteristic is a technique used to evaluate the prediction ability of a binary classifier (Zweig and Campbell, 1993). In the context of MID classifiers, sensitivity is defined as the probability of a classifier predicting conflict correctly and specificity is the probability of a classifier predicting peace correctly (Westin, 2001). The ROC curve is then a graph, which plots the sensitivity on the vertical-axis, and 1-specificity on the horizontal-axis, which is, also called false positive rate. The area under curve (AUC) is used as a measure to compare the performance of each classifier. The AUC for NN and SVM are 0.81 and 0.84 with standard errors of 0.00998 and 0.01022, respectively. According to Hanley and McNeil (1983), the normal distribution z value, which is used to compare if there is a significant difference between AUCs of two classifiers that are derived from the same cases, is given by:

$$z = \frac{A_1 - A_2}{\sqrt{SE_1^2 + SE_2^2 - 2rSE_1SE_2}} \qquad (13)$$

where *$A_1$, $A_2$,, $SE_2$* and *$SE_2$* are the areas and standard errors of the respective curves. The value *r* represents the estimated correlation between $A_1$ and $A_2$ (Hanley and McNeil, 1983). The value of *z* is 2.697, which gives significant difference in a 95% confidence interval. The results of the SVM are much better in predicting the conflicts without affecting the prediction of peace as it is clearly shown in I. The ROC graphs of the NN and SVM results are given in figure 3.

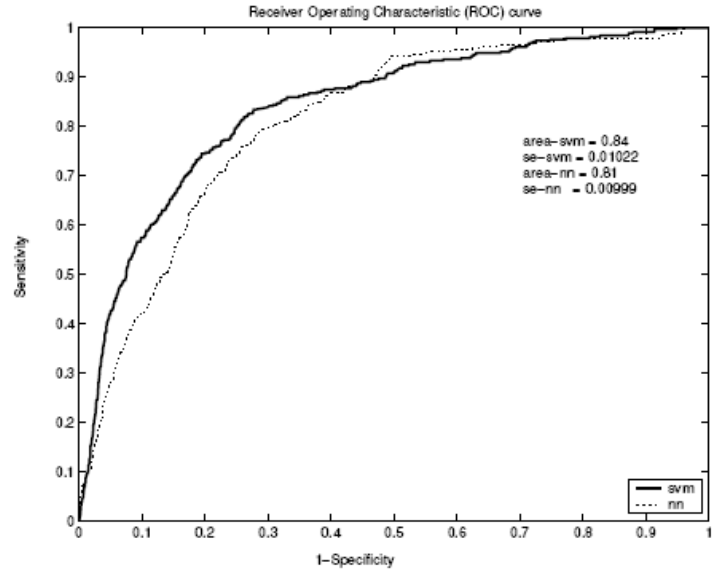

Fig. 3. ROC curve for both NN and SVM. Area-svm and area-nn signify the areas under the curves while se-svm and se-nn are their respective standard errors.

*B. Influence of each variable on the MID outcome*

In order to see the influence of each variable on the MID result, two separate sensitivity analysis were done for NN and SVM. The two techniques agree in picking up the influences of some of the variables while they differ on others.

 1*) Experiment one*: This experiment looked at how assigning each variable to its possible maximum value while keeping the rest at their possible minimum values and vise versa affect the MID outcome. The results for NN show that *democracy level* and *capability ratio* are able to deliver a peaceful outcome while all the other variables are kept minimal. This means, dyadic preponderance has a deterring effect to conflict, as is joint democracy of the states involved. On the other hand, keeping all the variables to their maximum values while assigning one variable to its minimum value resulted in a peaceful outcome. In other words, no single variable is able to change the outcome if all the other variables are set to their possible maximum values for NN. A similar experiment conducted for SVM shows that it is not able to pick the influence of a variable using the same approach, as it is possible with NN. That is, whether the variables are set to their minimum or maximum gives a peace outcome.

2) *Experiment two*: This experiment was done to measure the sensitivity of the variables in the spirit of partial derivatives as (Zeng, 1999) has put it. The idea is basically to see the change in the output for a change in one of the input variables. The experiment looks at how the MID varies when one variable is assigned to its possible maximum and minimum values while keeping all the other variables constant. The results found for both NN and SVM are shown in II. Our test data set has 26737 cases of peace and 392 cases of war. The first line of the table shows the correct number of peace and war prediction when all variables are used. Assigning each variable to take its possible maximum and minimum values while keeping the other variables fixed then generated different testing data sets. Each subsequent line of the table depicts the number of correct prediction of peace and war.

**NN result**: It shows *democracy level* has the maximum effect in reducing conflict while *capability ratio* is second in conformance to the first experiment. Allowing *democracy* to have its possible maximum value for the whole data set was able to avoid conflict totally. *Capability ratio* reduced the occurrence of conflict by 98%. Maximising alliance between the dyads reduced the number of conflicts by 20%. Maximising *dependency* has a 6% effect in reducing possible conflicts. Reducing *major power* was able to cut the number of conflicts by 3%. Minimising the *contiguity* of the dyads to their possible lower values and maximising the *distance* reduced the number of conflicts by 45% and 31%, respectively. This last result agrees with the realist theory that says far apart countries have less reasons to have conflicts (Oneal and Russett, 1999).

TABLE II
THE EFFECT OF CHANGING ONE VARIABLE WHILE KEEPING THE OTHER VARIABLES FIXED

| Variable | NN Peace | NN War | SVM Peace | SVM War |
|---|---|---|---|---|
| Test set results | 19464 | 297 | 20914 | 295 |
| Dem-min | 16263 | 325 | 22327 | 205 |
| Dem-max | 26345 | - | 23761 | 35 |
| Allies-min | 18555 | 313 | 20469 | 274 |
| Allies-max | 21034 | 237 | 21999 | 153 |
| Contig-min | 23682 | 164 | 24745 | 60 |
| Contig-max | 12463 | 342 | 18939 | 281 |
| Dist-min | 5351 | 370 | 25067 | 34 |
| Dist-max | 22525 | 206 | 26284 | 3 |
| Capab-min | 6929 | 373 | 19840 | 180 |
| Capab-max | 26322 | 3 | 26345 | - |
| Depnd-min | 19455 | 297 | 20498 | 305 |
| Depnd-max | 20411 | 277 | 26345 | - |
| Majpow-min | 19686 | 289 | 22345 | - |
| Majpow-max | 19428 | 299 | 23583 | 136 |

**SVM result:** The results of the experiment show inconsistency on how the MID outcome is affected when the variables are maximised and minimised. Further investigation is required to understand more clearly the influence of each variable (e.g. exploring some other sensitivity analysis techniques). Therefore, an alternative sensitivity analysis that involves using only one explanatory variable to predict the MID and see the goodness of accuracy is used. The ROC curves were drawn and the area under the curve (AUC) calculated for the purpose of ranking as is suggested in (Guyon and Elisseeff, 2003). The rankings of the effects of variables on the MID for NN and SVM vary as shown in table III. The reason for the variance may be because in the case of NN the chain effect of changes in one variable to the other variables is accounted for (as the variables are believed to be highly interdependent (Beck, King, and Zeng, 2000) as opposed to the case of SVM where the effect of one variable is considered separately.

TABLE III
RANKINGS OF THE INFLUENCE OF VARIABLES FOR NN AND SVM

| Rank | NN | SVM |
| --- | --- | --- |
| 1 | Democracy | Contiguity |
| 2 | Capability | Distance |
| 3 | Contiguity | Major power |
| 4 | Distance | Capability |
| 5 | Alliance | Democracy |
| 6 | Dependency | Dependency |
| 7 | Major power | Alliance |

V. CONCLUSION

In this paper two artificial intelligence techniques, neural networks (NN) and support vector machine (SVM), are used to predict militarised interstate disputes. The independent/input variables are *Democracy*, *Allies*, *Contingency*, *Distance*, *Capability*, *Dependency* and *Major Power* while the dependent/ output variable is *MID* result which is either peace or conflict. A neural network trained with scaled conjugate gradient algorithm and an SVM with a radial basis kernel function together with grid-search and cross-validation techniques were employed to find the optimal model.

The results found show that SVM has better capacity in forecasting conflicts without effectively affecting the correct peace prediction than NN. Two separate experiments were conducted to see the influence of each variable to the MID outcome. The first one assigns each variable to its possible highest value while keeping the rest to their possible lowest values. The NN results show that both *democracy level* and *capability ratio* are able to influence the outcome to be peace. On the other hand, none of the variables was able to influence the MID outcome to be conflict when all the other variables were maximum. SVM was not able to pick the effects of the variable for this experiment.

The second experiment assigns each variable to its possible highest or lowest value while keeping the other variables fixed to their original values. The results agree with the previous experiment. If we group the variables in terms of their effect and rank them, Democracy level and capability ratio are first, contiguity, distance and alliance second and dependency, major power are ranked third using NN. Although SVM performs better than NN, the results of NN are easier to be interpreted in relation to variable influence.

Biographies

Eyasu Hayemariam is an MSc in Electrical Engineering student at the University of the Witwatersrand in South Africa. He graduated with a BSc Degree in Statistics at the University of Osmara in Eritrea

Tshilidzi Marwala received a BS in Mechanical Engineering at Case Western Reserve University in Ohio, an MSc in Mechanical Engineering from University of Pretoria and a PhD in Artificial Intelligence from the University of Cambridge. Previously he was a post-doctoral fellow at Imperial College (London). He is currently a professor at the University of the Witwatersrand in South Africa.

Monica Lagazio holds a PhD in Politics and Artificial Intelligence from Nottingham University and an MA in Politics from the University of London. Before joining the University of Kent at Canterbury in 2004, she was Lecturer in Politics at the University of the Witwatersrand and Research Fellow at Yale University. She also held a position of senior consultant in the economic and financial service of one of the leading global consulting companies in London.